\documentclass[11pt]{article}

\usepackage[final]{acl}

\usepackage{times}
\usepackage{latexsym}
\usepackage{amssymb}

\usepackage[T1]{fontenc}

\usepackage[utf8]{inputenc}
\usepackage{microtype}

\usepackage{inconsolata}

\usepackage{graphicx}
\usepackage{booktabs}
\usepackage{multirow}
\usepackage{xcolor}
\usepackage{colortbl}
\usepackage{amsmath}

\title{Rethinking Visual Neglect: Steering via Context-Preference for MLLM Hallucination Mitigation}

\author{
  \textbf{Jingwen Wu}\textsuperscript{1}, 
  \textbf{Xijun Zhang}\textsuperscript{1}, 
  \textbf{Ge Song}\textsuperscript{1}\thanks{~Corresponding author.} \\
  \textsuperscript{1}School of Computer and Electronic Information, Nanjing Normal University, China \\
  \texttt{jingwen.wu@19230415.edu.cn}
}

\begin{document}
\maketitle

\begin{abstract}
Object hallucination remains a primary obstacle to the reliable deployment of Multimodal Large Language Models (MLLMs). Current inference-time mitigation methods mainly assume hallucinations stem from \textit{visual neglect}, steering models to enhance visual reliance. In contrast, our systematic interventions on multiple MLLMs show that pushing toward more visual reliance may \textit{exacerbate} hallucinations on some models, while less may \textit{mitigates} hallucinations. This result suggests that attributing hallucinations solely to visual insufficiency is underdetermined. We argue that the image, as a context, simultaneously competes with the model's parametric knowledge and the textual context. For this, we propose a training-free framework, \textbf{Context-Preference Activation Steering} (CAS). It extracts two semantically distinct \textbf{Context Preference Vectors} (CPVs) via two small sets of designed conflict samples and applies them via single-pass signed residual injection at mid-early MLP layers during inference to control information reliance. Experiments show that CAS substantially mitigates object hallucinations without increasing decoding latency and preserves native text-generation quality.
\end{abstract}

\begin{figure}[t]
\centering
\includegraphics[width=0.95\linewidth,trim=5 0 0 5, clip]{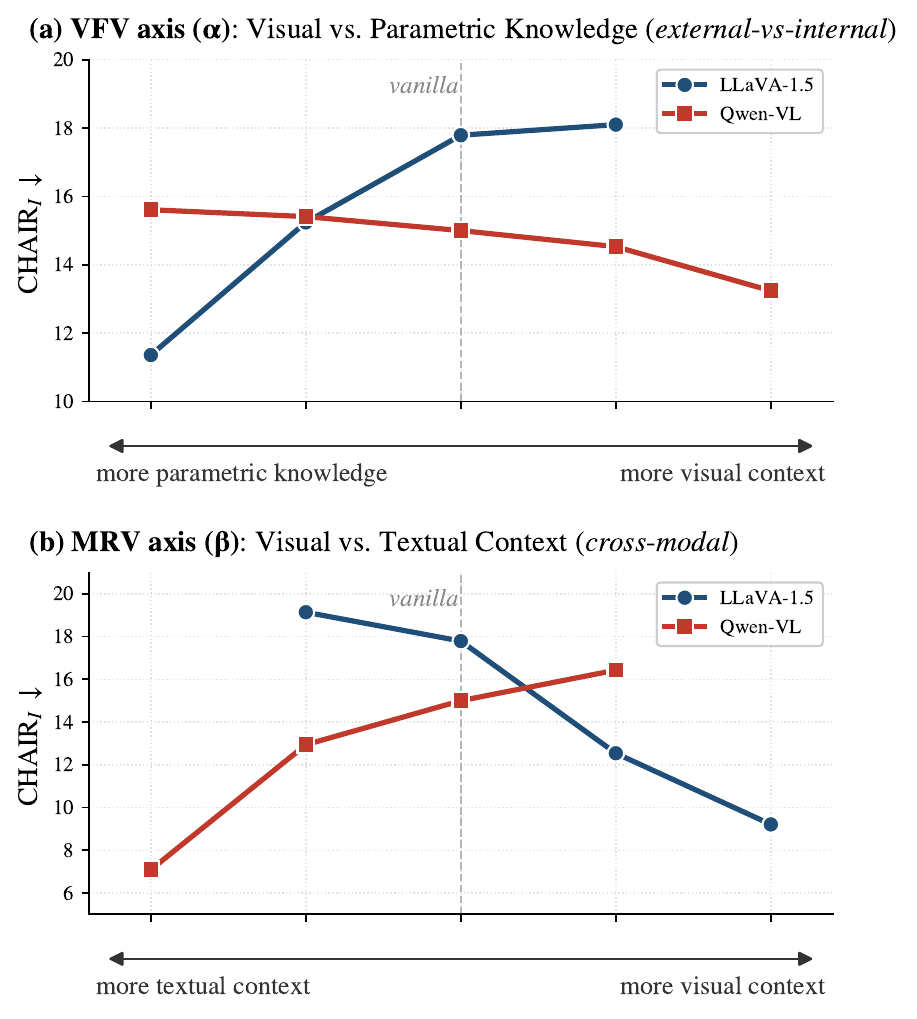}
\caption{\textbf{CHAIR$_I$ of LLaVA-1.5 and Qwen-VL under VFV and MRV interventions.} This reveals three phenomena: (i) steering toward \textit{more image context} on either CPV may instead exacerbate hallucinations; (ii) the response slopes across the two CPVs can take opposite signs within the same model; (iii) on the same CPV, different models exhibit divergent trends.}
\label{fig:hook}
\end{figure}

\section{Introduction}
\label{sec:intro}
Multimodal Large Language Models (MLLMs) have made rapid progress in visual understanding and open-ended generation, yet object hallucination, in which the model describes objects absent from the input image, remains a core obstacle to reliable deployment~\cite{li2023pope}. The problem is particularly severe in long-form generation (e.g., detailed captioning): as the generation length grows, the model progressively drifts away from visual evidence and hallucinations accumulate~\cite{zhang2025longer, zhou2024lure}. In the literature, given the prohibitive cost of retraining and alignment fine-tuning, training-free inference-time intervention has emerged as an attractive route.

However, current mainstream inference-time mitigation methods (e.g., PAI~\cite{zheng2024pai}, AttnReal~\cite{tu2025attnreal}, VCD~\cite{leng2024vcd}) almost rest on the same implicit assumption that hallucinations stem from the model's \textit{visual neglect} and its over-reliance on textual priors. They therefore encourage the model to \textit{pay more attention to the image} through various interventions, regardless of the base model. Yet the gains of these methods are inconsistent across base models, suggesting that \textit{visual neglect} alone is an incomplete characterization of the reason for MLLM hallucination.

This paper offers a systematic re-examination of this consensus. We argue that the image input, as a context, simultaneously competes against two semantic signals inside the model: the model's parametric knowledge from pretraining and the textual context formed jointly by the textual instruction and the generation history. However, prior work has not examined how these two competitions jointly shape hallucinations. Based on this characterization, we decouple visual preference along two distinct competing dimensions: the \textit{external-vs-internal} competition, and the \textit{cross-modal} competition between visual and textual context. We accordingly construct two conflict-based sample sets: \textit{counterfactual images} that violate commonsense and \textit{symmetric image-text conflict pairs}. By capturing the model's hidden states under these conflicting inputs, we extract two semantically distinct \textbf{Context Preference Vectors (CPVs)}: the \textbf{Visual Fidelity Vector (VFV)}, which captures the \textit{external-vs-internal} preference, and the \textbf{Modality Reliance Vector (MRV)}, which captures the \textit{cross-modal} preference (see \S\ref{subsec:cpv_extraction}). 

Figure~\ref{fig:hook} reports the results of using the two CPVs to perform visual-dependency interventions on LLaVA-1.5 and Qwen-VL. We observe that: (i) intervening toward \textit{more visual context} does not always reduce hallucinations, as witnessed by the two ascending curves in the figure; (ii) on the same model, the optimal sign of intervention on VFV and MRV can be opposite, indicating strong independence between the two behavioral responses; (iii) for the same CPV, the sign of the positive response differs across models. These phenomena reveal not only the incompleteness of the simplistic \textit{more visual, less hallucination} assumption, but also the heterogeneity of different models with respect to different context preferences.

Based on the above observation, we can conclude that each model has, on both VFV and MRV, an effective sign that substantially reduces hallucinations. Therefore, we propose \textbf{Context-Preference Activation Steering (CAS)}, a training-free inference-time framework. CAS requires only a small set of task-irrelevant samples to extract VFV and MRV within minutes. In inference, it injects signed residuals of the CPVs at the MLP outputs of just four mid-layers, adapting the intervention to the specific model's visual-preference response. Experiments on mainstream base models and benchmarks show that CAS substantially mitigates hallucinations in a single forward pass without extra decoding latency and preserving native generation quality.

The main contributions are summarized as follows:
\begin{itemize}
    \item \textbf{A decoupled characterization of MLLM visual preference.} We identify an incompleteness in prior work's characterization of visual preference, decouple it into two semantically distinct Context Preference Vectors (CPVs), namely VFV (\textit{external-vs-internal}) and MRV (\textit{cross-modal}), and experiments reveal the intrinsic heterogeneity of different models in visual-preference responses and the behavioral independence between the two CPVs.
    \item \textbf{The CAS framework.} We propose \textbf{Context-Preference Activation Steering (CAS)}, a training-free inference-time intervention framework. CAS first extracts the two CPVs from a small set of conflict-based samples by reading candidate-answer preferences in the model's hidden states. In inference, it injects signed residuals of the CPVs at intermediate MLLM layers. CAS is light plug-and-play and can mitigate hallucinations without degrading text quality.
    \item \textbf{Systematic empirical evaluation and analysis.} Extensive experiments on different base models and benchmarks verify that CAS efficiently mitigates MLLM hallucinations while avoiding the repetitive degeneration of some decoding-based intervention methods. We further find that different MLLM layer bands respond markedly differently to the two CPVs. This offers a new analytical perspective on how multimodal models process visual-linguistic semantic conflicts.
\end{itemize}

\section{Related Work}
\label{sec:related_work}
\subsection{Training-based Hallucination Mitigation}
\label{subsec:training_based}
A straightforward approach to mitigating MLLM hallucinations is to construct hallucination-related preference data and perform alignment training on the model. DPO-based methods form preferences between \textit{faithful} and \textit{hallucinated} responses to suppress hallucinations. Recent object-aware and perception-enhanced DPO variants further extend this line by incorporating hallucination-aware preference construction and multimodal perception priors~\cite{compagnoni2025chairdpo, feng2026peadpo}. For example, Re-Align~\cite{yu2025realign} constructs dual text-visual preference pairs via image retrieval and fine-tunes VLMs with retrieval-augmented DPO; RLHF-V~\cite{yu2024rlhfv} brings RLHF ideas into MLLM hallucination mitigation by collecting segment-level fine-grained correctional feedback and performing behavioral alignment via Dense DPO. These methods yield stable gains but require large preference datasets and substantial training resources. Besides, aligned models sometimes underperform the original VLM on general tasks~\cite{yu2025realign}, and the coarse-grained preferences of standard RLHF are prone to \textit{reward hacking} and \textit{behavior degeneration}~\cite{yu2024rlhfv}. Consequently, even with substantial human correction and computational resources, training-based methods struggle to suppress hallucinations without compromising general capability.

\subsection{Training-free Mitigation}
\label{subsec:training_free}

Beyond alignment training, a complementary family of methods mitigates hallucinations without modifying the base VLM's parameters. We review three sub-classes below.

\subsubsection{Detect-then-Correct Workflows}
\label{subsec:workflow}

These methods construct detect-then-correct post-hoc workflows that perform multi-stage review and rewriting of generated outputs. LURE~\cite{zhou2024lure} trains a hallucination revisor that takes the original caption, localizes hallucinated objects, and rewrites them into faithful descriptions. Woodpecker~\cite{yin2024woodpecker} further decomposes detection and correction into five stages, sequentially invoking external VLMs to complete sub-tasks such as key concept extraction, visual claim verification, and text rewriting. Recent generate-then-verify variants adopt retrospective verification and resampling to reduce unsupported generations~\cite{wu2025generateverify}. These methods are plug-and-play, but each sample requires 3-5 sequential model calls or external VLMs and detectors, causing inference latency and computational overhead.

\subsubsection{Retrospection and Layer-Contrastive Decoding}
\label{subsec:retrospection}
These methods assume that the model already encodes implicit signals of \textit{relative faithfulness}, which need only be detected and exploited during decoding to suppress hallucinations. OPERA~\cite{huang2024opera} detects an 'over-trust' attention pattern during beam search and backtracks to reselect tokens. DoLa~\cite{chuang2024dola} and its MLLM extension DeCo~\cite{xia2024deco} ground the same signal at the layer level by contrasting the vocabulary-projected logits of higher layers against those of earlier layers, retaining tokens with higher cross-layer consistency. Recent work further improves contrastive decoding robustness via decoupled objectives and attention-steerable contrastive signals~\cite{chen2025dcd, wang2025ascd,chuang2024dola}. However, retrospection requires maintaining multiple beam backtracks, which disrupts streaming output and incurs inference overhead. The layer-contrastive decoding exacerbates repetitive degeneration and is sensitive to the choice of layers.

\subsubsection{Methods under the Visual-Neglect Hypothesis}
\label{subsec:visual_neglect}

Within this family, work builds on the hypothesis that hallucinations stem from \textit{visual neglect} and forces \textit{pay more attention to the image} by enhancing reliance on visual signals or suppressing textual priors. On the \textit{decoding side}, VCD~\cite{leng2024vcd} contrasts the output logits of the original image against those of a Gaussian-noised image to suppress residual linguistic priors that persist after visual degradation. On the \textit{forward-pass side}, PAI~\cite{zheng2024pai} amplifies the attention weights of image tokens, complemented by contrastive logits from text-only input. AttnReal~\cite{tu2025attnreal} redistributes the excessive attention accumulated on historical tokens back to visual tokens. DMAS~\cite{yin2026dmas} injects dynamically retrieved steering vectors into deep-layer attention heads; SSL~\cite{chen2025ssl} leverages a pretrained sparse autoencoder to apply predefined faithful and hallucination feature vectors to visual and output tokens; V-ITI~\cite{li2025viti} trains per-head linear classifiers to detect \textit{visual neglect} and selectively inject visual signals. \cite{li2025cai, zhu2026lookcarefully} extend this line with caption-sensitive attention intervention and adaptive visual reinforcement strategies.

Despite the success of these methods, there remain some limitations: (i) the assumption that \textit{amplifying visual signals suffices to reduce hallucinations} overlooks the behavioral heterogeneity across different base models, resulting in limited hallucination reduction on some models; (ii) substantially suppressing hallucinations may disrupt the model's native generation manifold and lead to repetitive degeneration, as shown in our experiments (\S\ref{subsec:gen_results}); (iii) the extraction of steering signals of activation-steering methods relies heavily on external models or pretrained resources.

Unlike previous methods, CAS introduces a decoupled view of visual preference and injects two signed CPVs, being flexible to align with different base models' natural responses. The CPVs can be extracted within minutes and only require a small amount of context-conflict data. CAS's intervention is complete in a single forward pass and targets only four middle MLP layers, achieving high efficiency with minimal disruption to the model's native generation manifold.

\section{Methodology}
\label{sec:method}
\begin{figure*}[t]
\centering
\includegraphics[width=\linewidth,trim=0 10 5 10, clip]{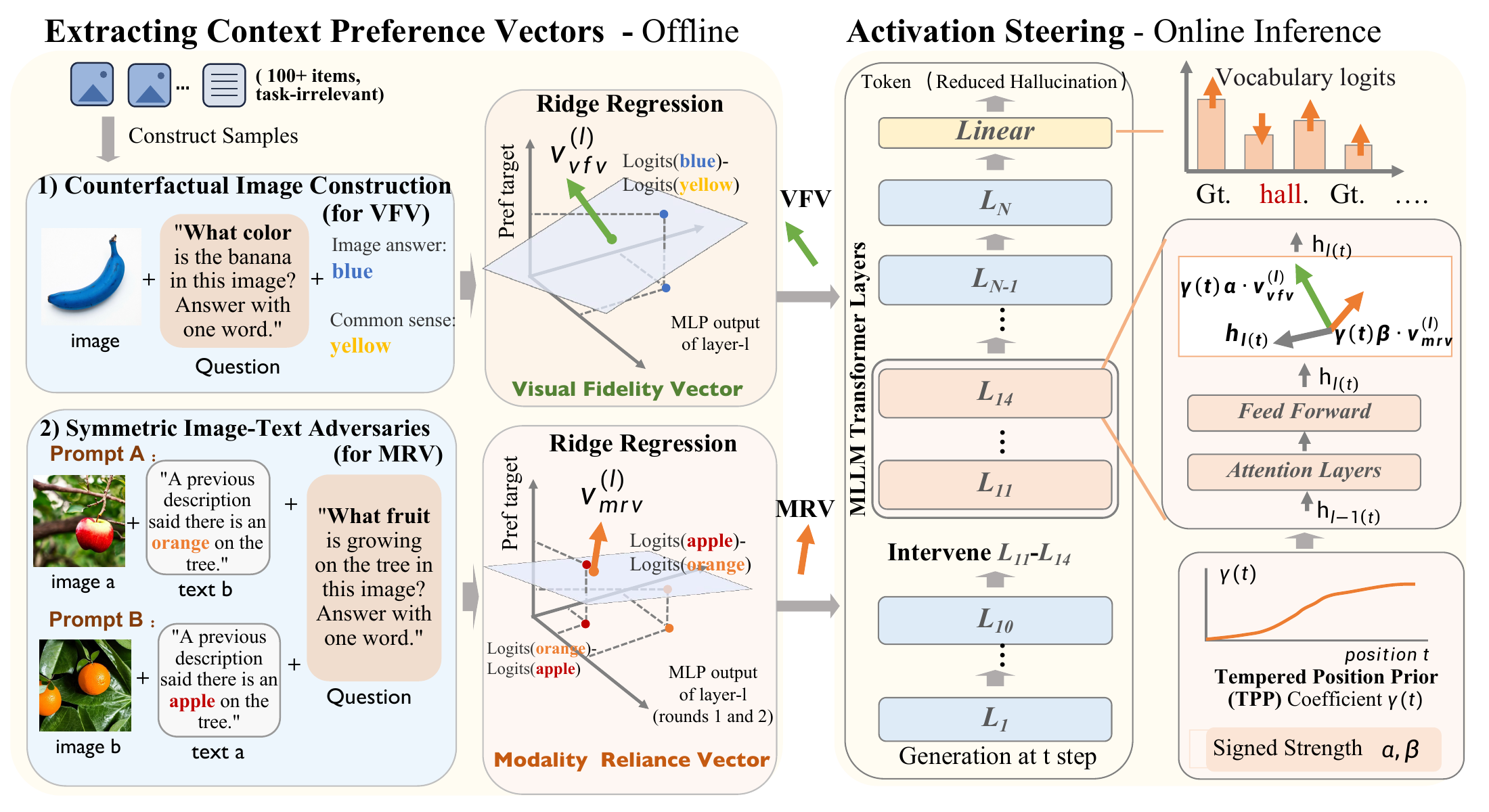}
\caption{\textbf{Overview of the CAS framework.}
\textbf{Left: Extracting Context Preference Vectors (offline).} Construct two types of context-conflict setups, and read the layer-$l$ MLP output together with the first-token logit difference $\mathrm{pref}$, and obtain $v_{\text{vfv}}^{(l)}$ (VFV) and $v_{\text{mrv}}^{(l)}$ (MRV) via layer-wise ridge regression.
\textbf{Right: Activation Steering (online inference).} At inference, we apply an additive residual $\gamma(t)\cdot(\alpha\, v_{\text{vfv}}^{(l)} + \beta\, v_{\text{mrv}}^{(l)})$ only at the MLP outputs of the mid-early layers L11--14, where $\alpha, \beta$ are signed-strength coefficients and $\gamma(t)$ is the position-wise gate given by the Tempered Position Prior.}
\label{fig:cas_method}
\end{figure*}

We propose \textbf{Context-Preference Activation Steering (CAS)} (Figure~\ref{fig:cas_method}), a training-free framework for mitigating MLLM hallucinations. CAS consists of two components: (i) \textbf{Extracting Context Preference Vectors} (§\ref{subsec:cpv_extraction}); (ii) \textbf{Signed Activation Steering} (§\ref{subsec:signed_steering}).

\subsection{Extracting Context Preference Vectors}
\label{subsec:cpv_extraction}
We decouple visual preference into two axes: the visual context with the parametric knowledge and with the textual context, i.e., \textit{external-vs-internal} and \textit{cross-modal}.

\paragraph{Setup~I: Counterfactual Image Construction.} To extract the \textbf{Visual Fidelity Vector (VFV)} that captures the \textit{visual context vs.\ parametric knowledge} preference, we construct a set of counterfactual images that violate commonsense. Each sample takes the form $(I_{\mathrm{cf}}, q, y^+, y^-)$, where $I_{\mathrm{cf}}$ is the counterfactual image, $q$ is the question, $y^+$ corresponds to the image content, and $y^-$ corresponds to commonsense knowledge. Thus, the model's choice over $(y^+, y^-)$ directly reflects its relative preference between the visual context and parametric knowledge.

\paragraph{Setup~II: Symmetric Image-Text Conflict Pairs.} To extract the \textbf{Modality Reliance Vector (MRV)} that captures the \textit{visual context vs.\ textual context} preference, we first select a pair of equally weighted but semantically distinct candidate answers $(A, B)$ (e.g., \textit{apple} vs.\ \textit{orange}; \textit{blue} vs.\ \textit{green}). To eliminate the interference caused by any prior preference of the model toward one specific answer (i.e., the residual influence of parametric knowledge), we construct two symmetric image-text subsets:
\begin{gather}
\mathcal{S}_a = \{(I_A, q_B, y^+=A, y^-=B)\}, \\
\mathcal{S}_b = \{(I_B, q_A, y^+=B, y^-=A)\},
\end{gather}
where $I_X$ denotes an image whose content is $X$, and $q_X$ denotes a text prompt whose content is $X$. Since the answers favoring visual evidence and those favoring the textual prior are exactly flipped between $\mathcal{S}_a$ and $\mathcal{S}_b$, biases tied to specific token frequencies cancel out when the two halves are jointly fitted. This symmetric design ensures that the extracted MRV captures the cross-modal preference rather than vocabulary-frequency noise.

The detailed data construction of the two sample sets is provided in Appendix~\ref{app:cpv_construction}.

\paragraph{Context Preference Vectors Extraction.}
The two setups share the same extraction procedure. We choose the MLP layers as the extraction site, since the MLP has been shown as the key module in the Transformer for storing factual knowledge~\cite{geva2021transformer} and for causally determining outputs~\cite{meng2022locating}. For each sample, we let the MLLM autoregressively generate the answer, locate the earliest occurrence of $y^+$ or $y^-$ within the generated token sequence, and record the position of its first token as $t^\star$. At the preceding position $t^\star{-}1$, we read the next-token logits $\ell \in \mathbb{R}^{|V|}$ that predict the token at $t^\star$, and define the continuous preference signal.
\begin{equation}
\mathrm{pref} = \max_{i \in \mathcal{T}^+} \ell_i - \max_{j \in \mathcal{T}^-} \ell_j,
\end{equation}
where $\mathcal{T}^+$ is the set of first-token candidates for the answer $y^+$ aligned with the visual context, and $\mathcal{T}^-$ is the set of first-token candidates for the answer $y^-$ aligned with the opposing context (parametric knowledge or textual context), with case and leading-space variants included. At the same position $t^* - 1$, we also read the MLP output $h_l \in \mathbb{R}^d$ of every layer $l$ (i.e., the hidden state of that layer) and perform a per-layer $\ell_2$-regularized ridge regression
\begin{equation}
\min_{w_l, b_l} \sum_{s} \big( \mathrm{pref}_s - w_l^\top \, h_l^{(s)} - b_l \big)^2 + \lambda \|w_l\|^2,
\end{equation}
where samples whose behavioral signal is weak ($|\mathrm{pref}| < \epsilon$) are uniformly removed before fitting. The weight vector $w_l \in \mathbb{R}^d$ obtained independently at each layer yields the corresponding Context Preference Vector: the weight fitted on Setup~I is denoted $v_{\text{vfv}}^{(l)}$ (VFV); the weight jointly fitted over the two halves $\mathcal{S}_a \cup \mathcal{S}_b$ of Setup~II is denoted $v_{\text{mrv}}^{(l)}$ (MRV).s

\paragraph{Layer localization.} We fix both CPV extraction and residual injection to the mid-early MLP layers L11-14; this band is determined by the layer-wise attribution experiments reported in §\ref{subsec:analysis}.

\subsection{Signed Activation Steering}
\label{subsec:signed_steering}
At inference, CAS formalizes the intervention as a \textbf{single-pass signed residual injection} on the MLP outputs of the target layers. For a target layer $l \in \{11,12,13,14\}$ and a generation step $t$,
\begin{equation}
\hat{h}_l^{(t)} \;=\; h_l^{(t)} \,+\, \gamma(t)\!\cdot\!\big(\alpha\, v_{\text{vfv}}^{(l)} \,+\, \beta\, v_{\text{mrv}}^{(l)}\big),
\end{equation}
where $\alpha$ and $\beta$ are the \textbf{signed} strength coefficients of VFV and MRV, respectively, and $\gamma(t)$ is the position-wise temporal gate. During prefill, the intervention is applied to the last prompt token. During decoding, it is triggered once per newly generated token, adding the residual above to the MLP output of the target layer.

\paragraph{Tempered Position Prior (TPP)} In long-form generation, the hallucination rate varies markedly with decoding time, i.e., later positions are increasingly prone to drifting away from the visual evidence~\cite{zhang2025longer, zhou2024lure}. To achieve a better trade-off between intervention strength and native generation quality, we couple the intervention strength to a position-level hallucination probability, ensuring high-risk positions with strong interventions. Specifically, we estimate the prior hallucination probability of each position on a separate calibration set by calculating the fraction of hallucinated object tokens generated by the base model at each position. Because the per-token statistic exhibits high variance and adjacent positions are intrinsically correlated, we apply variable-length position bucketing (details in Appendix~\ref{app:impl}) to aggregate neighboring positions and suppress random noise, yielding a per-bucket hallucination fraction $P_b$. Then, we apply a power transform with temperature $T$ to obtain the bucket coefficient:
\begin{equation}
c_b \;=\; \big(P_b \,/\, \max_{b'} P_{b'}\big)^{1/T},\qquad b\in\mathcal{B}.
\end{equation}
$T>1$ raises the low-bucket coefficients (making buckets more uniform); $T=1$ recovers the original distribution. Generative tasks take $\gamma(t) = c_{b(t)}$ as the dynamic temporal gate, while discriminative tasks, which output only a single token without autoregressive drift, set $\gamma(t) \equiv 1$. To isolate the independent impacts of steering directions, we enforce $\gamma(t) \equiv 1$ across all ablation studies. The position dependence of $c_b$ is fit on the calibration set. Calibration-set size, bucket boundaries, and other details are listed in Appendix~\ref{app:impl}.

\section{Experiments}
\label{sec:experiments}

\subsection{Experimental Setup}
\label{subsec:setup}

\textbf{Models.} We evaluate CAS on four open-source MLLMs: LLaVA-1.5~\cite{liu2024llava15}, Shikra~\cite{chen2023shikra}, Qwen-VL~\cite{bai2023qwenvl}, and InstructBLIP~\cite{dai2023instructblip}. These models differ in cross-modal fusion design, allowing us to assess the cross-architecture applicability of CAS.

\textbf{Benchmarks.} We cover both generative and discriminative hallucination evaluations. Generative: (1) \textbf{CHAIR}~\cite{rohrbach2018chair}, on 500 COCO~\cite{lin2014coco} images for detailed captioning, reporting CHAIR$_S$, CHAIR$_I$, and F1; (2) \textbf{AMBER}~\cite{wang2024amber} generative split (1{,}004 images), reporting Hal, CHAIR, Cog, and Cover. Discriminative: (3) \textbf{POPE}~\cite{li2023pope}, with 3{,}000 questions in each of the random / popular / adversarial subsets, reporting Acc and F1.

\textbf{Baselines.} We compare vanilla greedy decoding against eight training-free methods, ordered by publication date: DoLa~\cite{chuang2024dola}, VCD~\cite{leng2024vcd}, OPERA~\cite{huang2024opera}, PAI~\cite{zheng2024pai}, Code~\cite{kim2024code}, DeCo~\cite{xia2024deco}, AttnReal~\cite{tu2025attnreal}, and SSL~\cite{chen2025ssl}. All methods follow their original recommended configurations.

\textbf{Text-quality metrics.} Some baselines exhibit repetitive generation (representative cases in Appendix~\ref{app:case_studies}). Such repetition is not penalized by CHAIR / Hal / F1; worse, because CHAIR$_I$ is counted at the token-instance level without deduplication, repeated GT-grounded objects deflate its denominator. We therefore additionally report \textbf{Rep} ($n$-gram repetition rate) for generative tasks. Definitions are provided in Appendix~\ref{app:impl}.

\textbf{CAS configuration.} CAS performs a single-pass signed residual injection on the MLP outputs of layers L11-L14 for each base model. Full hyperparameter configurations are listed in Appendix~\ref{app:config}.

\textbf{Decoding \& hardware.} All methods use greedy decoding to eliminate sampling variance, with \texttt{max\_new\_tokens=512} for CHAIR / AMBER and 64 for POPE. All experiments are conducted on a single NVIDIA RTX 3090. Prompt templates for each benchmark are listed in Appendix~\ref{app:impl}.

\subsection{Generative Hallucination}
\label{subsec:gen_results}

\begin{table*}[t]
\centering
\small
\resizebox{\textwidth}{!}{%
\begin{tabular}{l | ccc | ccc | ccc | ccc}
\toprule
\multirow{2}{*}{Method} & \multicolumn{3}{c|}{LLaVA-1.5} & \multicolumn{3}{c|}{Shikra} & \multicolumn{3}{c|}{Qwen-VL} & \multicolumn{3}{c}{InstructBLIP} \\
& CHAIR$_S\downarrow$ & CHAIR$_I\downarrow$ & Rep & CHAIR$_S\downarrow$ & CHAIR$_I\downarrow$ & Rep & CHAIR$_S\downarrow$ & CHAIR$_I\downarrow$ & Rep & CHAIR$_S\downarrow$ & CHAIR$_I\downarrow$ & Rep \\
\midrule
Vanilla   & 57.6 & 17.79 & 1.7 & 61.0 & 18.58 & 0.4 & 50.8 & 15.00 & 0.6 & 52.6 & 16.83 & 4.4 \\
DoLa      & 74.4 & 27.23 & \textcolor{red}{9.4} & 50.6 & 16.78 & \textcolor{red}{285.7} & 53.6 & 16.71 & 0.2 & 73.0 & 33.54 & \textcolor{red}{38.3} \\
VCD       & 57.6 & 18.82 & 2.0 & 64.0 & 19.49 & 0.2 & 51.0 & 14.99 & 0.5 & 54.8 & 17.44 & 4.8 \\
OPERA     & 70.4 & 22.21 & \textcolor{red}{66.3} & 57.4 & 21.37 & \textcolor{red}{17.7} & 52.0 & 21.62 & 0.0 & 59.6 & 20.29 & 0.2 \\
PAI       & \underline{37.4} & \textbf{10.13} & \textcolor{red}{27.2} & \underline{49.6} & \underline{14.84} & \textcolor{red}{1.7} & -- & -- & -- & -- & -- & -- \\
Code      & 51.8 & 16.90 & 1.1 & 62.2 & 19.24 & 0.9 & 53.6 & 16.84 & 0.6 & 52.2 & 17.41 & 3.6 \\
DeCo      & 46.2 & 14.10 & 5.0 & 54.0 & 16.06 & 0.6 & \underline{45.4} & 13.16 & 1.1 & \underline{45.4} & \underline{14.49} & 9.8 \\
AttnReal  & 45.4 & 13.97 & \textcolor{red}{8.4} & 60.4 & 17.34 & 1.1 & 49.0 & \underline{13.30} & 1.6 & 47.8 & 14.93 & \textcolor{red}{19.2} \\
SSL       & 44.8 & 15.53 & \textcolor{red}{7.6} & -- & -- & -- & -- & -- & -- & 48.0 & 15.74 & 5.6 \\
\rowcolor{gray!10}
\textbf{CAS } & \textbf{35.8} & \underline{12.21} & 0.8 & \textbf{44.0} & \textbf{13.56} & 0.6 & \textbf{38.0} & \textbf{12.19} & 1.4 & \textbf{36.0} & \textbf{13.01} & 3.8 \\
\bottomrule
\end{tabular}%
}
\caption{CHAIR results. CHAIR$_S$ / CHAIR$_I$ lower is better. Rep ($\times 10^{-3}$); Rep $\ge 3\times$ Vanilla highlighted in \textcolor{red}{red}. Best CHAIR$_S$ / CHAIR$_I$ per model in \textbf{bold}, second best \underline{underlined}. ``--'' indicates the method is not available for that base model. Full results including F1 and TTR are reported in Appendix~\ref{app:chair_full} (Table~\ref{tab:chair_full}).}
\label{tab:chair}
\end{table*}

\begin{table}[t]
\centering
\small
\resizebox{\columnwidth}{!}{%
\begin{tabular}{llccccc}
\toprule
Model & Method & Hal$\downarrow$ & CHAIR$\downarrow$ & Cog$\downarrow$ & Cover$\uparrow$ & Rep \\
\midrule
\multirow{6}{*}{LLaVA-1.5}
  & Vanilla  & 31.1 & 7.7 & 3.6 & 49.4 & 0.5 \\
  & VCD      & 36.2 & 8.7 & 3.9 & \underline{50.4} & 0.5 \\
  & PAI      & \underline{23.2} & \textbf{4.4} & \textbf{1.6} & 50.5 & \textcolor{red}{29.0} \\
  & AttnReal & 26.0 & \underline{5.3} & \underline{2.4} & 50.2 & \textcolor{red}{4.9} \\
  & SSL      & 31.6 & 8.3 & 3.4 & 44.2 & 0.4 \\
  & CAS      & \textbf{17.2} & \textbf{4.4} & \textbf{1.6} & 46.7 & 0.1 \\
\midrule
\multirow{6}{*}{Shikra}
  & Vanilla  & 48.5 & 10.8 & 5.5 & 51.4 & 0.3 \\
  & VCD      & 50.0 & 10.7 & 4.9 & 52.3 & 0.2 \\
  & PAI      & \textbf{21.4} & \underline{8.2} & \textbf{1.9} & \textcolor{red}{33.5} & \textcolor{red}{1.4} \\
  & AttnReal & 43.3 & 8.7 & 4.3 & \underline{51.7} & 0.5 \\
  & CAS      & \underline{34.2} & \textbf{7.6} & \underline{2.9} & 50.5 & 0.1 \\
\midrule
\multirow{5}{*}{Qwen-VL}
  & Vanilla  & 29.6 & \underline{6.3} & 2.6 & \underline{53.3} & 0.2 \\
  & VCD      & 37.5 & 8.8 & 3.4 & 50.8 & 0.2 \\
  & AttnReal & \underline{25.9} & \textbf{5.3} & \underline{2.1} & 54.2 & \textcolor{red}{0.7} \\
  & CAS      & \textbf{23.5} & \textbf{5.3} & \textbf{1.9} & 50.2 & 0.5 \\
\midrule
\multirow{4}{*}{InstructBLIP}
  & Vanilla  & 35.3 & 8.1 & 3.7 & \underline{53.8} & 2.3 \\
  & VCD      & 39.1 & 8.3 & 3.7 & 54.2 & 2.1 \\
  & AttnReal & \underline{30.5} & \textbf{5.7} & \underline{2.9} & 54.3 & \textcolor{red}{27.2} \\
  & CAS      & \textbf{28.8} & \underline{6.1} & \textbf{2.8} & 51.9 & 2.6 \\
\bottomrule
\end{tabular}%
}
\caption{AMBER generative results. Hal / CHAIR / Cog lower is better ($\downarrow$); Cover higher is better ($\uparrow$). Rep ($\times 10^{-3}$); Rep values $\ge 3\times$ Vanilla and abnormal Cover collapses are highlighted in \textcolor{red}{red}. Cover / Rep are reported as reference and not bolded; \textbf{bold} marks best per column among the three hallucination metrics (Hal / CHAIR / Cog), second best \underline{underlined}.}
\label{tab:amber}
\end{table}

CHAIR results are shown in Table~\ref{tab:chair}. CAS attains the lowest or tied-lowest CHAIR$_S$ on all base models, reducing CHAIR$_S$ relative to vanilla by 37.8\% / 27.9\% / 25.2\% / 31.6\% on LLaVA-1.5 / Shikra / Qwen-VL / InstructBLIP, respectively. DoLa, OPERA, PAI, AttnReal, and SSL exhibit red-flagged Rep values on multiple base models, indicating visible repetitive degeneration, whereas CAS's Rep stays on the order of vanilla. Full results (including F1 and TTR) are reported in Appendix~\ref{app:chair_full}, where CAS's F1 is no lower than vanilla on any base model; length-aligned CHAIR results are reported in Appendix~\ref{app:lenalign}.

AMBER results are shown in Table~\ref{tab:amber}. To balance evaluation cost with coverage, we include a curated subset of competitive baselines on AMBER. CAS attains the lowest Hal on LLaVA, Qwen, and InstructBLIP, with Cover maintained near vanilla; relative to vanilla, CAS reduces Hal by 44.7\% / 20.6\% / 18.4\% on the three models. On Shikra, PAI attains a lower Hal but its Cover collapses by 34.8\% (51.4 $\rightarrow$ 33.5, red); CAS instead reduces Hal by 29.5\% while preserving Cover.

\subsection{Discriminative Hallucination}
\label{subsec:pope_results}

\begin{table}[t]
\centering
\small
\caption{POPE results (averaged over random / popular / adversarial subsets). Best per column within each model in \textbf{bold}, second best \underline{underlined}. ``--'' indicates the method is not available for that base model. Per-subset breakdown in Appendix~\ref{app:pope_full} (Table~\ref{tab:pope_full}).}
\label{tab:pope}
\resizebox{=0.85\columnwidth}{!}{
\begin{tabular}{l | cc | cc}
\toprule
& \multicolumn{2}{c|}{LLaVA-1.5} & \multicolumn{2}{c}{Shikra} \\
\cmidrule(lr){2-3} \cmidrule(lr){4-5}
Method & Acc$\uparrow$ & F1$\uparrow$ & Acc$\uparrow$ & F1$\uparrow$ \\
\midrule
Greedy & 85.17 & 83.49 & 77.21 & 80.23 \\
DoLa   & 77.78 & 75.12 & 63.30 & 70.28 \\
VCD    & 84.37 & 83.22 & 74.13 & 78.46 \\
OPERA  & \underline{85.26} & 83.63 & 76.98 & 79.42 \\
PAI    & 85.17 & \underline{83.83} & 78.37 & 75.26 \\
Code   & 85.04 & 83.29 & \underline{78.50} & \underline{80.82} \\
DeCo   & 85.17 & 83.49 & 70.44 & 76.18 \\
SSL    & 84.94 & 83.11 & -- & -- \\
\rowcolor{gray!10}
\textbf{CAS (ours)} & \textbf{86.58} & \textbf{86.11} & \textbf{78.57} & \textbf{81.20} \\
\midrule
\midrule
& \multicolumn{2}{c|}{Qwen-VL} & \multicolumn{2}{c}{InstructBLIP} \\
\cmidrule(lr){2-3} \cmidrule(lr){4-5}
Method & Acc$\uparrow$ & F1$\uparrow$ & Acc$\uparrow$ & F1$\uparrow$ \\
\midrule
Greedy & 87.67 & 87.10 & 76.42 & 79.51 \\
DoLa   & 81.83 & 84.20 & 56.16 & 60.87 \\
VCD    & 87.33 & \underline{87.20} & 75.08 & 78.14 \\
OPERA  & 86.25 & 87.13 & 69.23 & 72.90 \\
Code   & 87.31 & 87.12 & 77.06 & 79.72 \\
DeCo   & \underline{87.65} & 87.12 & 76.80 & 79.69 \\
SSL    & -- & -- & \underline{78.77} & \underline{81.02} \\
\rowcolor{gray!10}
\textbf{CAS (ours)} & \textbf{87.84} & \textbf{87.53} & \textbf{84.69} & \textbf{84.74} \\
\bottomrule
\end{tabular}
}
\end{table}

POPE results are shown in Table~\ref{tab:pope} (averaged over the three subsets; per-subset breakdown in Appendix Table~\ref{tab:pope_full}). CAS attains the top average Acc and F1 on all four models, improving F1 on InstructBLIP by 6.6\% relative to vanilla. In the per-subset results, PAI's F1 on Shikra drops by 6.5\% / 4.4\% relative to vanilla on the popular / adversarial subsets, whereas CAS is no lower than vanilla on any (base model, subset) pair.

\subsection{Per-model response to VFV and MRV}
\label{subsec:per_cpv_response}

\begin{figure}[t]
\centering
\includegraphics[width=\linewidth,trim=5 5 0 5, clip]{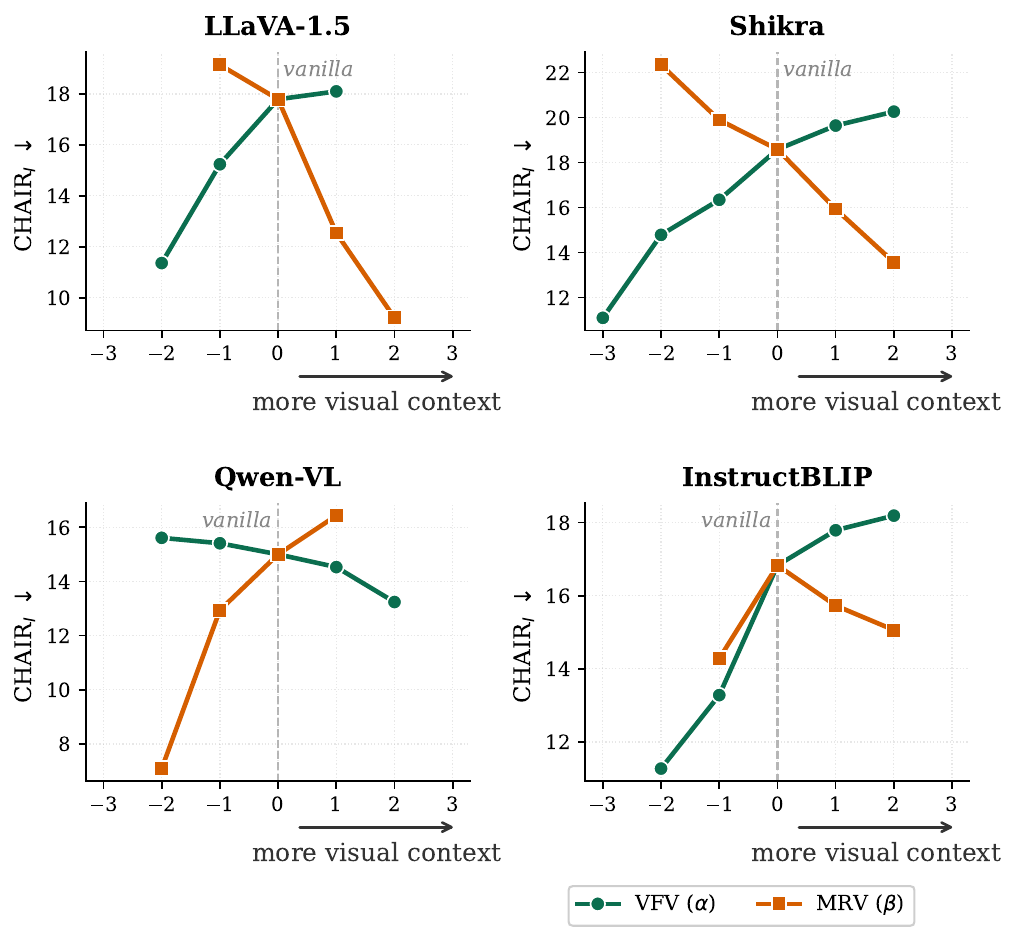}
\caption{\textbf{CHAIR$_I$ under VFV-only and MRV-only intervention.} Each subplot corresponds to one base model and reports CHAIR$_I$ across signed intensities $\alpha$ (VFV) and $\beta$ (MRV); the dashed line marks that model's vanilla CHAIR$_I$. Full numbers are in Appendix Tables~\ref{tab:sweep_mrv} and~\ref{tab:sweep_vfv}.}
\label{fig:sweep_4models}
\end{figure}
The intensity coefficients $\alpha$ and $\beta$ of CAS are two signed scalars. Figure~\ref{fig:sweep_4models} reports VFV-only and MRV-only sweeps on the four base models (without positional priors). Models respond to VFV and MRV in markedly different ways: the four models already exhibit three distinct response patterns. More visual emphasis does not automatically lower hallucination; nevertheless, for every model both VFV and MRV admit signed settings that bring CHAIR$_I$ below vanilla, and some models are particularly sensitive to one of the two. CAS exploits this per-model response to substantially reduce hallucination (full numbers in Appendix Tables~\ref{tab:sweep_mrv} and~\ref{tab:sweep_vfv}).

\paragraph{VFV and MRV are geometrically independent.} Since both VFV and MRV concern vision, we further investigate whether they are two estimates of the same underlying preference. We compute $\cos(v_{\text{vfv}}^{(l)},\, v_{\text{mrv}}^{(l)})$ on the intervention layers (L11--14) of four base models. All 16 values fall within $[-0.06,\, +0.11]$, with mean $|\cos| \approx 0.033$, matching the order of magnitude expected of two random vectors in $\mathbb{R}^{4096}$. VFV and MRV are therefore approximately orthogonal in the latent space, corresponding to two independent control dimensions.

\subsection{Layer Analysis}
\label{subsec:analysis}

To verify the choice of intervention layers, we place the same CPVs into different four-layer windows on LLaVA-CHAIR. To amplify the differences between windows, we use stronger fixed intensities (MRV $\beta{=}{+}2$, VFV $\alpha{=}{-}2$). As shown in Table~\ref{tab:layer}, the two CPVs follow a consistent pattern across windows: within the windows we sweep, the mid-early band L11--14 is the only band that is both effective and stable; from L16 onward the effect returns to the vanilla level and is essentially null; at the shallow layers L1--4 and L6--9, both MRV and VFV trigger severe degeneration and diverge into runaway repetition. We therefore fix the intervention at L11--14 in all experiments.

\begin{table}[t]
\centering
\small
\begin{tabular}{llcc}
\toprule
Window & CPV & CHAIR$_S\downarrow$ & CHAIR$_I\downarrow$ \\
\midrule
\multirow{2}{*}{L1--4 }
  & MRV & \multicolumn{2}{c}{\textcolor{red}{\textit{degenerate, Rep $\approx 575\times$ }}} \\
  & VFV & \multicolumn{2}{c}{\textcolor{red}{\textit{degenerate, Rep $\approx 587\times$ }}} \\
\midrule
\multirow{2}{*}{L6--9}
  & MRV & \multicolumn{2}{c}{\textcolor{red}{\textit{degenerate, Rep $\approx 278\times$ }}} \\
  & VFV & \multicolumn{2}{c}{\textcolor{red}{\textit{degenerate, Rep $\approx 280\times$ }}} \\
\midrule
\multirow{2}{*}{\textbf{L11--14} (selected)}
  & MRV & \textbf{18.8} & \textbf{9.21}  \\
  & VFV & \textbf{35.4} & \textbf{11.36} \\
\midrule
\multirow{2}{*}{L16--19}
  & MRV & 58.8 & 18.35 \\
  & VFV & 56.4 & 18.04 \\
\midrule
\multirow{2}{*}{L21--24}
  & MRV & 56.8 & 17.33 \\
  & VFV & 55.8 & 17.75 \\
\midrule
\multirow{2}{*}{L26--29}
  & MRV & 55.8 & 18.19 \\
  & VFV & 57.2 & 18.23 \\
\midrule
\multicolumn{2}{l}{\textit{vanilla}} & 57.6 & 17.79 \\
\bottomrule
\end{tabular}
\caption{\textbf{Layer ablation.} Effect of MRV ($\beta{=}{+}2$) and VFV ($\alpha{=}{-}2$) interventions across different four-layer windows on LLaVA-CHAIR. CHAIR$_S$ / CHAIR$_I$ lower is better. Degenerate rows are marked in the CHAIR columns as ``Rep $N\times$'', where $N$ is the Rep multiplier relative to vanilla. The selected window \textbf{L11--14} is in bold.}
\label{tab:layer}
\end{table}

\section{Conclusion}
\label{sec:conclusion}
This paper challenges the prevailing assumption that MLLM object hallucinations stem solely from visual neglect, demonstrating instead that increasing visual reliance can unexpectedly exacerbate hallucinations due to a three-way competition among image context, internal parametric knowledge, and textual context. By decoupling this phenomenon into two distinct dimensions, the Visual Fidelity Vector (VFV) and the Modality Reliance Vector (MRV), we introduce Context-Preference Activation Steering (CAS), a lightweight training-free framework that precisely guides internal information reliance during inference. By injecting the signed residuals of these vectors into mid-early MLP layers in a single forward pass, CAS uniquely adapts to each model's idiosyncratic architectural preferences. Extensive evaluations across four prominent MLLMs and three benchmarks validate that CAS significantly suppresses hallucinations without increasing decoding latency or degrading native text quality. We hope that treating visual preference as a per-model control signal rather than a uniform \textit{more-visual} intervention provides a useful perspective for inference-time MLLM mitigation.

\section*{Limitations}
Although the Context-Preference Activation Steering (CAS) framework proposed in this study demonstrates significant empirical effectiveness in mitigating hallucinations in Multimodal Large Language Models (MLLMs), several limitations remain. First, our current research scope primarily focuses on object hallucination; the effectiveness of the CAS framework in addressing attribute misdescriptions, spatial and logical relationship misjudgments, and more complex multimodal reasoning hallucinations has yet to be empirically verified. Second, constrained by computational resources, all base models evaluated in this study are limited to the 7B parameter scale. The framework's effectiveness on larger models remains to be explored, and the optimal intervention window currently identified at the mid-early MLP layers may experience positional drift within deeper network architectures. Furthermore, during the extraction of Context Preference Vectors (CPVs), there are certain compromises in the generation quality and fidelity of the samples used to construct conflict pairs, which might introduce minor visual or textual biases into the extracted vectors. Finally, while we have empirically validated the efficacy of our approach, we have not yet conducted an in-depth theoretical analysis of its underlying mechanisms, specifically regarding how it interacts with multimodal feature fusion pathways of multimodal features within the model's deep representation space.

\bibliography{custom}

\appendix

\section{Inference Latency}
\label{app:latency}

We measure the per-token decoding latency of each method on LLaVA-1.5; results are reported in Table~\ref{tab:latency}. CAS performs only an additive residual on the MLP outputs of four mid-early layers within a single forward pass, with no additional forward pass and no contrastive decoding. Its per-token latency is therefore close to vanilla and substantially lower than double-forward methods (PAI / VCD / CODE) and multi-step search methods (OPERA).

\begin{table}[h]
\centering
\small
\begin{tabular}{lcc}
\toprule
Method & Per-token Latency (ms) & std \\
\midrule
Vanilla   & 29.65 & $\pm$2.76 \\
\rowcolor{gray!10}
\textbf{CAS (ours)} & \textbf{28.09} & $\pm$1.05 \\
SSL       & 28.49 & $\pm$1.05 \\
DoLa      & 33.68 & $\pm$1.40 \\
AttnReal  & 35.08 & $\pm$2.56 \\
DeCo      & 40.97 & $\pm$1.22 \\
PAI       & 56.35 & $\pm$1.33 \\
VCD       & 56.04 & $\pm$1.10 \\
CODE      & 113.05 & $\pm$12.33 \\
OPERA     & 464.03 & $\pm$385.38 \\
\bottomrule
\end{tabular}
\caption{Per-token decoding latency (ms) on LLaVA-1.5 setting. The only extra cost of CAS comes from one additive residual at the mid-early MLP outputs, of the same order as vanilla.}
\label{tab:latency}
\end{table}

\section{Per-CPV Sweeps: MRV-only and VFV-only on Four Models}
\label{app:per_cpv_sweep}

For each of the four base models, we sweep MRV-only ($\alpha{=}0$, varying $\beta$) and VFV-only ($\beta{=}0$, varying $\alpha$). Full results are reported in Tables~\ref{tab:sweep_mrv} and~\ref{tab:sweep_vfv}. Three observations follow. (i) For each base model, both VFV and MRV admit signed settings that bring CHAIR$_S$ below vanilla. (ii) The beneficial sign combination $(\alpha, \beta)$ varies across base models; at least three distinct sign patterns therefore coexist across the four base models. (iii) Injecting either CPV at extreme intensities disrupts the model's native generation manifold and yields the degenerate rows; nevertheless, for every base model there exists an intensity at which the CPV substantially reduces hallucination without disrupting the manifold, and CAS operates within this regime.

\begin{table}[h]
\centering
\small
\begin{tabular}{llccc}
\toprule
Model & $\beta$ & CHAIR$_S\downarrow$ & CHAIR$_I\downarrow$ & Rep \\
\midrule
\multirow{5}{*}{LLaVA-1.5}
  & $-2$ & \multicolumn{2}{c}{\textit{degenerate}} & \textcolor{red}{828.2} \\
  & $-1$ & 67.8 & 19.14 & \textcolor{red}{12.7} \\
  & $\phantom{+}0$ & 57.6 & 17.79 & 1.7 \\
  & $+1$ & 36.6 & 12.54 & 1.0 \\
  & $+2$ & 18.8 & \phantom{1}9.21 & 2.6 \\
\midrule
\multirow{5}{*}{Shikra}
  & $-2$ & 68.9 & 22.34 & \textcolor{red}{1.6} \\
  & $-1$ & 65.8 & 19.89 & 0.6 \\
  & $\phantom{+}0$ & 61.0 & 18.58 & 0.4 \\
  & $+1$ & 55.2 & 15.93 & 0.3 \\
  & $+2$ & 44.0 & 13.56 & 0.7 \\
\midrule
\multirow{5}{*}{Qwen-VL}
  & $-2$ & 21.0 & \phantom{1}7.09 & \textcolor{red}{2.5} \\
  & $-1$ & 41.0 & 12.94 & 1.1 \\
  & $\phantom{+}0$ & 50.8 & 15.00 & 0.6 \\
  & $+1$ & 53.8 & 16.43 & 0.7 \\
  & $+2$ & \multicolumn{2}{c}{\textit{degenerate}} & \textcolor{red}{29.8} \\
\midrule
\multirow{5}{*}{InstructBLIP}
  & $-2$ & \multicolumn{2}{c}{\textit{degenerate}} & \textcolor{red}{110.1} \\
  & $-1$ & 38.4 & 14.28 & 12.6 \\
  & $\phantom{+}0$ & 52.6 & 16.83 & \phantom{1}4.4 \\
  & $+1$ & 49.4 & 15.73 & \phantom{1}3.8 \\
  & $+2$ & 50.0 & 15.06 & \phantom{1}5.8 \\
\bottomrule
\end{tabular}
\caption{\textbf{MRV-only sweep} ($\alpha{=}0$, CHAIR, no position prior). Rep is reported in units of $10^{-3}$; values $\ge 3\times$ the vanilla row ($\beta{=}0$) of each model are highlighted in red.}
\label{tab:sweep_mrv}
\end{table}

\begin{table}[h]
\centering
\small
\begin{tabular}{llccc}
\toprule
Model & $\alpha$ & CHAIR$_S\downarrow$ & CHAIR$_I\downarrow$ & Rep \\
\midrule
\multirow{5}{*}{LLaVA-1.5}
  & $+2$ & \multicolumn{2}{c}{\textit{degenerate}} & \textcolor{red}{121.3} \\
  & $+1$ & 53.0 & 18.10 & 1.5 \\
  & $\phantom{+}0$ & 57.6 & 17.79 & 1.7 \\
  & $-1$ & 48.2 & 15.24 & 2.2 \\
  & $-2$ & 35.4 & 11.36 & \textcolor{red}{5.3} \\
\midrule
\multirow{6}{*}{Shikra}
  & $+2$ & 68.2 & 20.26 & 0.5 \\
  & $+1$ & 64.0 & 19.64 & 0.7 \\
  & $\phantom{+}0$ & 61.0 & 18.58 & 0.4 \\
  & $-1$ & 53.6 & 16.34 & 0.7 \\
  & $-2$ & 44.6 & 14.78 & 0.2 \\
  & $-3$ & 36.8 & 11.09 & 0.4 \\
\midrule
\multirow{6}{*}{Qwen-VL}
  & $+3$ & \multicolumn{2}{c}{\textit{degenerate}} & \textcolor{red}{10.6} \\
  & $+2$ & 41.8 & 13.24 & \textcolor{red}{2.7} \\
  & $+1$ & 48.2 & 14.53 & 1.2 \\
  & $\phantom{+}0$ & 50.8 & 15.00 & 0.6 \\
  & $-1$ & 50.6 & 15.41 & 0.6 \\
  & $-2$ & 49.8 & 15.61 & 0.8 \\
\midrule
\multirow{5}{*}{InstructBLIP}
  & $+2$ & 59.6 & 18.19 & 8.5 \\
  & $+1$ & 58.6 & 17.79 & 5.9 \\
  & $\phantom{+}0$ & 52.6 & 16.83 & 4.4 \\
  & $-1$ & 40.0 & 13.28 & 3.5 \\
  & $-2$ & 32.6 & 11.27 & 5.7 \\
\bottomrule
\end{tabular}
\caption{\textbf{VFV-only sweep} ($\beta{=}0$, CHAIR, no position prior). Rep is reported in units of $10^{-3}$; values $\ge 3\times$ the vanilla row ($\alpha{=}0$) of each model are highlighted in red.}
\label{tab:sweep_vfv}
\end{table}

\section{Length-aligned CHAIR}
\label{app:lenalign}

We truncate both CAS and vanilla captions to the first 64 tokens and recompute CHAIR. Table~\ref{tab:lenalign} shows that under this equal-budget setting, CAS's CHAIR$_S$ is substantially lower than vanilla on all four base models, while F1 is no lower than vanilla; this confirms that CAS's hallucination reduction also holds under shorter, length-aligned captions.

\begin{table}[h]
\centering
\small
\resizebox{\columnwidth}{!}{%
\begin{tabular}{llccc}
\toprule
Model & Method & CHAIR$_S \downarrow$ & CHAIR$_I \downarrow$ & F1 $\uparrow$ \\
\midrule
\multirow{2}{*}{LLaVA-1.5}
  & Vanilla & 43.4 & 13.51 & 74.76 \\
  & CAS     & \textbf{30.8} & \textbf{10.72} & \textbf{75.09} \\
\midrule
\multirow{2}{*}{Shikra}
  & Vanilla & 41.8 & 14.17 & 70.94 \\
  & CAS     & \textbf{32.0} & \textbf{10.43} & \textbf{74.08} \\
\midrule
\multirow{2}{*}{Qwen-VL}
  & Vanilla & 30.4 & \textbf{9.80} & 73.84 \\
  & CAS     & \textbf{28.4} & 9.95 & \textbf{74.78} \\
\midrule
\multirow{2}{*}{InstructBLIP}
  & Vanilla & 38.6 & 12.76 & 73.33 \\
  & CAS     & \textbf{28.2} & \textbf{9.97} & 73.31 \\
\bottomrule
\end{tabular}%
}
\caption{\textbf{Length-aligned CHAIR.} CHAIR$_S$ / CHAIR$_I$ (lower is better) and F1 (higher is better) recomputed after truncating all captions to the first 64 tokens. The best value within each model is in \textbf{bold}.}
\label{tab:lenalign}
\end{table}

\section{CPV Sample Construction}
\label{app:cpv_construction}

CAS extracts both CPVs from on the order of one hundred samples (Table~\ref{tab:cpv_counts}), far below the thousands-to-tens-of-thousands of prompts typically required by contrastive decoding or alignment fine-tuning. This data efficiency stems from the paradigm design rather than from any released dataset: context-conflict samples provide a high-SNR differential signal in the latent space (each sample carries its own $(y^+, y^-)$ candidate pair, and the symmetric image--text conflict pairs further cancel token-frequency bias via symmetric flipping), so ridge regression converges stably with on the order of one hundred samples. We do not claim that these samples constitute a high-quality dataset; on the contrary, as discussed in Limitations, the images of the symmetric image--text conflict pairs are generated by a 4\,GB-scale SD 1.5 model and are of limited quality. The core contribution is the paradigm of context-conflict samples combined with cross-symmetric subtraction, not the released samples themselves.

\begin{table}[ht]
\centering
\footnotesize
\caption{CPV sample counts. VFV is fitted alone on the counterfactual-image samples; MRV is jointly fitted on the two halves $\mathcal{S}_a \cup \mathcal{S}_b$ of the symmetric image--text conflict pairs, where the opposite-sign pref of the two halves automatically realizes cross-symmetric subtraction. Each sample carries a pair of first-token candidates $(y^+, y^-)$, which together define a continuous pref signal on that sample.}
\label{tab:cpv_counts}
\resizebox{\columnwidth}{!}{%
\begin{tabular}{llc}
\toprule
Setup & Form & Count \\
\midrule
Counterfactual image                  & $(I_{\text{cf}}, q, y^+, y^-)$ & 51 \\
Symmetric conflict pair $\mathcal{S}_a$ & $(I_A, q_B, y^+{=}A, y^-{=}B)$ & 55 \\
Symmetric conflict pair $\mathcal{S}_b$ & $(I_B, q_A, y^+{=}B, y^-{=}A)$ & 55 \\
\midrule
Total & & 161 \\
\bottomrule
\end{tabular}%
}
\end{table}

\paragraph{Prompts and candidate answers.} All $(q, y^+, y^-)$ triples (i.e., a question template paired with two candidate answers) are first drafted in a single LLM call and then verified manually. They cover more than ten commonsense concepts including color, material, shape, and counting; different samples within the same concept share the same question template with the specific object swapped, and the concept serves as the cross-validation grouping unit (samples of the same concept do not cross folds).

\paragraph{Counterfactual image generation.}
The counterfactual-image samples require images that violate commonsense (e.g., \textit{a blue banana}), which places a relatively high demand on the T2I model's text faithfulness: smaller T2I models tend to \textit{self-correct} the color back to the commonsense value. We therefore use the commercial-grade Qwen-Image-2.0 to generate the 51 images.

\paragraph{Symmetric image--text conflict pair generation.}
The symmetric image--text conflict pairs only require images of plain and unambiguous objects (e.g., \textit{pencil} vs.\ \textit{pen}), and the demand on T2I text faithfulness is low. We therefore use the open-source 4\,GB-scale Stable Diffusion v1.5 to generate the 110 images, trading off resource cost. This is also the MRV sample-quality constraint discussed in Limitations item~1: MRV may therefore still inherit some visual bias from SD 1.5. Re-extracting MRV with a stronger T2I model is a direct improvement target for future work.

\paragraph{Sample construction and CPV fitting time.}
Once the samples are ready, for each base model we run a forward pass over the 161 samples, read the hidden state at the focus token, and perform per-layer ridge regression. The entire pipeline finishes within a few minutes on a single NVIDIA RTX 3090. 

\section{Hyperparameter Configurations}
\label{app:config}

Table~\ref{tab:config} lists the hyperparameters per (base model, task type) combination: VFV intensity $\alpha$, MRV intensity $\beta$, and the position-prior temperature $T$ for generative tasks (no prior is used for discriminative tasks, marked ``--''). We find that the best $(\alpha, \beta)$ often differ between generative and discriminative tasks within the same base model; a plausible explanation is that the two task types place different demands on visual reliance: discriminative prompts explicitly name the queried object, giving the model a clear visual focus to locate, whereas generative captioning is free-form and has no such focus. The intervention layers are fixed at L11--14 across all base models.
\begin{table}[h]
\centering
\small
\resizebox{\columnwidth}{!}{%
\begin{tabular}{llccc}
\toprule
Model & Task type & $\alpha$ & $\beta$ & $T$ \\
\midrule
\multirow{2}{*}{LLaVA-1.5}
  & Generative      & $-2$ & $+1$ & 2 / 5 \\
  & Discriminative  & $+2$ & $+1$ & -- \\
\midrule
Shikra
  & All             & $0$  & $+2$ & -- \\
\midrule
\multirow{2}{*}{Qwen-VL}
  & Generative      & $+1$ & $-1$ & -- / 7 \\
  & Discriminative  & $+1$ & $+1$ & -- \\
\midrule
InstructBLIP
  & Generative      & $-2$ & $0$  & 1 \\
  & Discriminative  & $-2$ & $0$ & -- \\
\bottomrule
\end{tabular}%
}
\caption{CAS hyperparameters.}
\label{tab:config}
\end{table}

\section{CHAIR Full Results}
\label{app:chair_full}

Table~\ref{tab:chair_full} reports the full CHAIR results including F1 and TTR (the main paper Table~\ref{tab:chair} only reports CHAIR$_S$ / CHAIR$_I$ / Rep). CAS's F1 is no lower than vanilla across all four base models, indicating that the CHAIR reduction does not come from precision / recall trade-off; TTR remains close to vanilla, confirming that lexical diversity is preserved.

\begin{table*}[h]
\centering
\small
\resizebox{=0.9\textwidth}{!}{%
\begin{tabular}{l | ccccc | ccccc}
\toprule
& \multicolumn{5}{c|}{LLaVA-1.5} & \multicolumn{5}{c}{Shikra} \\
\cmidrule(lr){2-6} \cmidrule(lr){7-11}
Method & CHAIR$_S\downarrow$ & CHAIR$_I\downarrow$ & F1$\uparrow$ & Rep & TTR$\uparrow$ & CHAIR$_S\downarrow$ & CHAIR$_I\downarrow$ & F1$\uparrow$ & Rep & TTR$\uparrow$ \\
\midrule
Vanilla   & 57.6 & 17.79 & 75.2 & 1.7 & 0.703 & 61.0 & 18.58 & 71.8 & 0.4 & 0.732 \\
DoLa      & 74.4 & 27.23 & 68.3 & \textcolor{red}{9.4} & 0.747 & 50.6 & 16.78 & 68.6 & \textcolor{red}{285.7} & 0.489 \\
VCD       & 57.6 & 18.82 & 74.0 & 2.0 & 0.713 & 64.0 & 19.49 & 72.2 & 0.2 & 0.748 \\
OPERA     & 70.4 & 22.21 & 71.8 & \textcolor{red}{66.3} & 0.700 & 57.4 & 21.37 & 54.1 & \textcolor{red}{17.7} & 0.781 \\
PAI       & \underline{37.4} & \textbf{10.13} & 75.2 & \textcolor{red}{27.2} & 0.631 & \underline{49.6} & \underline{14.84} & 73.0 & \textcolor{red}{1.7} & 0.697 \\
Code      & 51.8 & 16.90 & 74.9 & 1.1 & 0.711 & 62.2 & 19.24 & 72.4 & 0.9 & 0.752 \\
DeCo      & 46.2 & 14.10 & 75.5 & 5.0 & 0.713 & 54.0 & 16.06 & \underline{74.5} & 0.6 & 0.737 \\
AttnReal  & 45.4 & 13.97 & \textbf{76.3} & \textcolor{red}{8.4} & 0.658 & 60.4 & 17.34 & 74.0 & 1.1 & 0.719 \\
SSL       & 44.8 & 15.53 & 74.1 & \textcolor{red}{7.6} & 0.693 & -- & -- & -- & -- & -- \\
\rowcolor{gray!10}
\textbf{CAS (ours)} & \textbf{35.8} & \underline{12.21} & \underline{76.0} & 0.8 & 0.703 & \textbf{44.0} & \textbf{13.56} & \textbf{75.4} & 0.6 & 0.733 \\
\midrule
\midrule
& \multicolumn{5}{c|}{Qwen-VL} & \multicolumn{5}{c}{InstructBLIP} \\
\cmidrule(lr){2-6} \cmidrule(lr){7-11}
Method & CHAIR$_S\downarrow$ & CHAIR$_I\downarrow$ & F1$\uparrow$ & Rep & TTR$\uparrow$ & CHAIR$_S\downarrow$ & CHAIR$_I\downarrow$ & F1$\uparrow$ & Rep & TTR$\uparrow$ \\
\midrule
Vanilla   & 50.8 & 15.00 & 75.4 & 0.6 & 0.728 & 52.6 & 16.83 & 72.9 & 4.4 & 0.687 \\
DoLa      & 53.6 & 16.71 & 72.5 & 0.2 & 0.746 & 73.0 & 33.54 & 64.8 & \textcolor{red}{38.3} & 0.680 \\
VCD       & 51.0 & 14.99 & 74.3 & 0.5 & 0.763 & 54.8 & 17.44 & 73.0 & 4.8 & 0.691 \\
OPERA     & 52.0 & 21.62 & 55.9 & 0.0 & 0.836 & 59.6 & 20.29 & 54.6 & 0.2 & 0.777 \\
Code      & 53.6 & 16.84 & 73.4 & 0.6 & 0.735 & 52.2 & 17.41 & 72.3 & 3.6 & 0.701 \\
DeCo      & \underline{45.4} & 13.16 & \underline{75.6} & 1.1 & 0.727 & \underline{45.4} & \underline{14.49} & 71.3 & 9.8 & 0.693 \\
AttnReal  & 49.0 & \underline{13.30} & \textbf{76.1} & 1.6 & 0.696 & 47.8 & 14.93 & \underline{73.9} & \textcolor{red}{19.2} & 0.643 \\
SSL       & -- & -- & -- & -- & -- & 48.0 & 15.74 & 73.1 & 5.6 & 0.672 \\
\rowcolor{gray!10}
\textbf{CAS (ours)} & \textbf{38.0} & \textbf{12.19} & \textbf{76.1} & 1.4 & 0.717 & \textbf{36.0} & \textbf{13.01} & \textbf{74.7} & 3.8 & 0.691 \\
\bottomrule
\end{tabular}%
}
\caption{CHAIR full results. CHAIR$_S$ / CHAIR$_I$ lower is better; F1 / TTR higher is better. Rep ($\times 10^{-3}$); Rep $\ge 3\times$ Vanilla highlighted in \textcolor{red}{red}; Rep / TTR are reported as reference only and not bolded. Best per column within each model in \textbf{bold}, second best \underline{underlined}.}
\label{tab:chair_full}
\end{table*}

\section{POPE Per-Subset Breakdown}
\label{app:pope_full}

Table~\ref{tab:pope_full} reports the full random / popular / adversarial breakdown of the POPE results that appear averaged in the main paper (Table~\ref{tab:pope}).

\begin{table*}[t]
\centering
\small
\resizebox{\textwidth}{!}{%
\begin{tabular}{l | cc cc cc | cc cc cc}
\toprule
& \multicolumn{6}{c|}{LLaVA-1.5} & \multicolumn{6}{c}{Shikra} \\
\cmidrule(lr){2-7} \cmidrule(lr){8-13}
& \multicolumn{2}{c}{Random} & \multicolumn{2}{c}{Popular} & \multicolumn{2}{c|}{Adversarial} & \multicolumn{2}{c}{Random} & \multicolumn{2}{c}{Popular} & \multicolumn{2}{c}{Adversarial} \\
\cmidrule(lr){2-3} \cmidrule(lr){4-5} \cmidrule(lr){6-7} \cmidrule(lr){8-9} \cmidrule(lr){10-11} \cmidrule(lr){12-13}
Method & Acc$\uparrow$ & F1$\uparrow$ & Acc$\uparrow$ & F1$\uparrow$ & Acc$\uparrow$ & F1$\uparrow$ & Acc$\uparrow$ & F1$\uparrow$ & Acc$\uparrow$ & F1$\uparrow$ & Acc$\uparrow$ & F1$\uparrow$ \\
\midrule
Greedy   & 86.57 & 84.80 & 85.40 & 83.69 & \textbf{83.53} & 81.98 & 80.73 & 82.65 & 77.97 & 80.85 & 72.93 & 77.18 \\
DoLa     & 82.10 & 78.32 & 75.90 & 73.74 & 75.33 & 73.29 & 65.43 & 71.57 & 62.97 & 70.35 & 61.50 & 68.91 \\
VCD      & 86.47 & 85.11 & 84.93 & 83.69 & 81.70 & 80.86 & 76.23 & 80.00 & 74.20 & 78.63 & 71.97 & 76.75 \\
OPERA    & 86.77 & 85.04 & 85.47 & 83.80 & \textbf{83.53} & \underline{82.04} & 79.73 & 81.36 & 78.27 & 80.30 & 72.93 & 76.60 \\
PAI      & \underline{87.07} & \underline{85.58} & \underline{85.70} & \underline{84.29} & 82.73 & 81.63 & 79.70 & 76.39 & \underline{78.73} & 75.57 & \textbf{76.67} & 73.81 \\
Code     & 86.43 & 84.59 & 85.30 & 83.51 & 83.40 & 81.77 & \underline{81.87} & \underline{83.37} & \textbf{78.83} & \underline{81.00} & \underline{74.80} & \underline{78.10} \\
DeCo     & 86.57 & 84.80 & 85.40 & 83.69 & \textbf{83.53} & 81.98 & 73.27 & 77.88 & 72.57 & 77.49 & 65.47 & 73.17 \\
SSL      & -- & -- & -- & -- & -- & -- & -- & -- & -- & -- & -- & -- \\
\rowcolor{gray!10}
\textbf{CAS (ours)} & \textbf{89.13} & \textbf{88.40} & \textbf{87.37} & \textbf{86.76} & 83.23 & \textbf{83.16} & \textbf{82.43} & \textbf{84.11} & 78.67 & \textbf{81.30} & 74.60 & \textbf{78.19} \\
\midrule
\midrule
& \multicolumn{6}{c|}{Qwen-VL} & \multicolumn{6}{c}{InstructBLIP} \\
\cmidrule(lr){2-7} \cmidrule(lr){8-13}
& \multicolumn{2}{c}{Random} & \multicolumn{2}{c}{Popular} & \multicolumn{2}{c|}{Adversarial} & \multicolumn{2}{c}{Random} & \multicolumn{2}{c}{Popular} & \multicolumn{2}{c}{Adversarial} \\
\cmidrule(lr){2-3} \cmidrule(lr){4-5} \cmidrule(lr){6-7} \cmidrule(lr){8-9} \cmidrule(lr){10-11} \cmidrule(lr){12-13}
Method & Acc$\uparrow$ & F1$\uparrow$ & Acc$\uparrow$ & F1$\uparrow$ & Acc$\uparrow$ & F1$\uparrow$ & Acc$\uparrow$ & F1$\uparrow$ & Acc$\uparrow$ & F1$\uparrow$ & Acc$\uparrow$ & F1$\uparrow$ \\
\midrule
Greedy   & 90.03 & 89.27 & 88.10 & 87.45 & \textbf{84.87} & \underline{84.57} & 78.23 & 80.75 & 78.00 & 80.58 & 73.03 & 77.19 \\
DoLa     & 88.27 & 89.05 & 82.23 & 84.30 & 75.00 & 79.24 & 51.57 & 58.43 & 58.23 & 61.97 & 58.67 & 62.22 \\
VCD      & \underline{90.37} & \underline{89.91} & 87.90 & 87.64 & 83.73 & 84.06 & 77.50 & 79.81 & 75.90 & 78.67 & 71.83 & 75.95 \\
OPERA    & 87.43 & 89.30 & 87.23 & 87.48 & 84.10 & \textbf{84.60} & 73.53 & 75.72 & 68.03 & 72.08 & 66.13 & 70.90 \\
PAI      & -- & -- & -- & -- & -- & -- & -- & -- & -- & -- & -- & -- \\
Code     & 90.27 & 89.76 & 88.03 & \underline{87.70} & 83.63 & 83.91 & 77.77 & 80.19 & 79.37 & 81.35 & 74.03 & 77.61 \\
DeCo     & 90.07 & 89.33 & \underline{88.17} & 87.55 & \underline{84.70} & 84.47 & 79.97 & 81.91 & 77.40 & 80.06 & 73.03 & 77.09 \\
SSL      & -- & -- & -- & -- & -- & -- & \underline{82.47} & \underline{83.73} & \underline{79.40} & \underline{81.41} & \underline{74.43} & \underline{77.92} \\
\rowcolor{gray!10}
\textbf{CAS (ours)} & \textbf{90.63} & \textbf{90.06} & \textbf{88.40} & \textbf{87.98} & 84.50 & 84.56 & \textbf{87.57} & \textbf{87.19} & \textbf{85.30} & \textbf{85.20} & \textbf{81.20} & \textbf{81.82} \\
\bottomrule
\end{tabular}%
}
\caption{POPE per-subset results. Best per column within each model in \textbf{bold}, second best \underline{underlined}. ``--'' indicates the method is not available for that base model.}
\label{tab:pope_full}
\end{table*}

\section{Implementation Details}
\label{app:impl}

\paragraph{CPV fitting.}
\begin{itemize}
\item Ridge regression with $\lambda = 1.0$, solved in closed form independently per layer.
\item 5-fold group-aware cross-validation, grouped by the commonsense concept each sample belongs to so that samples sharing a concept do not cross folds.
\item Samples with $|\mathrm{pref}| < 0.1$ are removed (signal too weak to constrain the regression).
\end{itemize}

\paragraph{Tempered Position Prior.}
\begin{itemize}
\item 13 buckets in total: one per 5 tokens in $[0, 20)$, one per 10 tokens in $[20, 100)$, and a single tail bucket for $\geq 100$.
\item Bucket coefficient $c_b = (P_b / \max_{b'} P_{b'})^{1/T}$, with $T$ from Table~\ref{tab:config}.
\item Calibration set: a COCO val2014 image subset disjoint from the CHAIR evaluation set. The base model generates captions on this set; we locate the token spans of hallucinated object words and compute the per-bucket fraction $P_b$. The $(\alpha, \beta)$ values in Table~\ref{tab:config} and the generative-task temperatures $T$ are selected on this same calibration set, which never touches any evaluation image.
\end{itemize}

\paragraph{Evaluation Prompts.} We adopt standard prompt templates tailored to each evaluation benchmark. For \textbf{CHAIR}, LLaVA-1.5, Shikra, and InstructBLIP use the standard captioning prompt: \texttt{"Please describe this image in detail."}. However, we observe that under this default instruction, Qwen-VL produces noticeably shorter captions than the other three base models. To prevent caption length from becoming a confounding factor in our CHAIR comparison, we apply a slightly shorter variant for Qwen-VL: \texttt{"Describe this image in detail."} , which successfully aligns its average generation length with the other models. For \textbf{AMBER generative}, we strictly adhere to the official evaluation protocol using \texttt{"Describe this image."}. For \textbf{POPE}, all models are evaluated under the standard binary classification template: \texttt{"Is there a [Object] in the image? Please answer yes or no."} , where \texttt{[Object]} is dynamically replaced per test item with the specific target object queried.

\paragraph{Text-Quality Metrics.}
To evaluate text quality and detect potential repetitive degeneration, we report the \textbf{Repetition rate (Rep).} and \textbf{Type-Token Ratio (TTR)}. Repetition is quantified via $\text{seq-rep-6}$, defined as $1 - |G_6^{\text{unique}}| / |G_6|$, where $G_6$ represents the multiset of all 6-grams in the generated caption. Lexical diversity is measured using the Moving-Average Type-Token Ratio (MATTR) with a fixed 50-token window slid one token at a time, which effectively mitigates the length-induced dilution effect inherent in standard TTR.

\section{Case Studies: Repetition Degeneration and CHAIR$_I$ Deflation}
\label{app:case_studies}

\begin{figure*}[h]
\centering
\includegraphics[width=\linewidth,trim=5 40 5 5, clip]{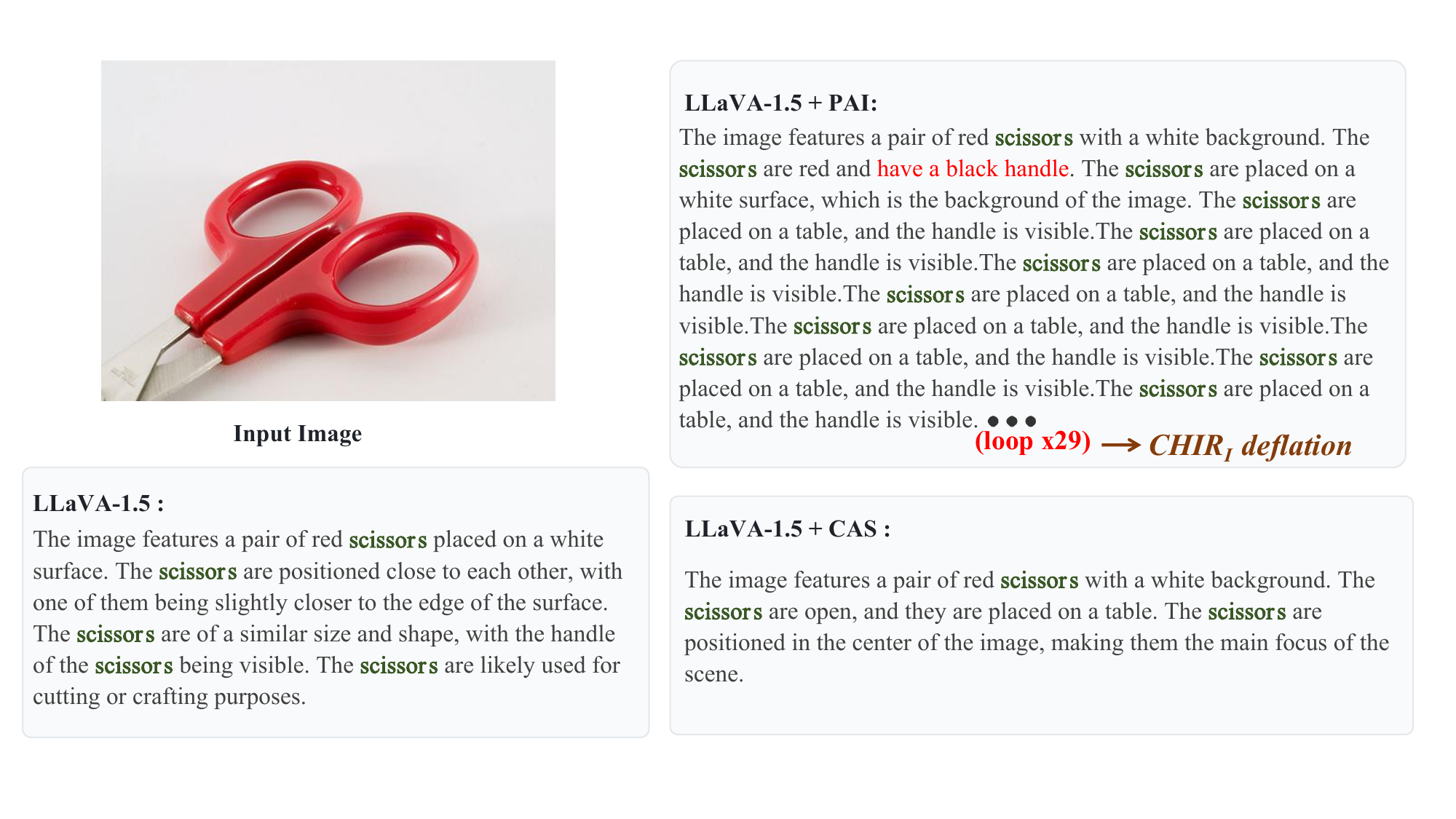}
\caption{\textbf{Repetitive degeneration of PAI on LLaVA-1.5.} A single hallucinated attribute (``black handle'') is repeated about 29 times, artificially inflating the mention count of GT-grounded objects; the token-level CHAIR$_I$ is therefore diluted by an inflated denominator and reads artificially low even though the caption has already degenerated. CAS produces a clean, repetition-free caption on the same image.}
\label{fig:case_pai}
\end{figure*}

\begin{figure*}[h]
\centering
\includegraphics[width=\linewidth]{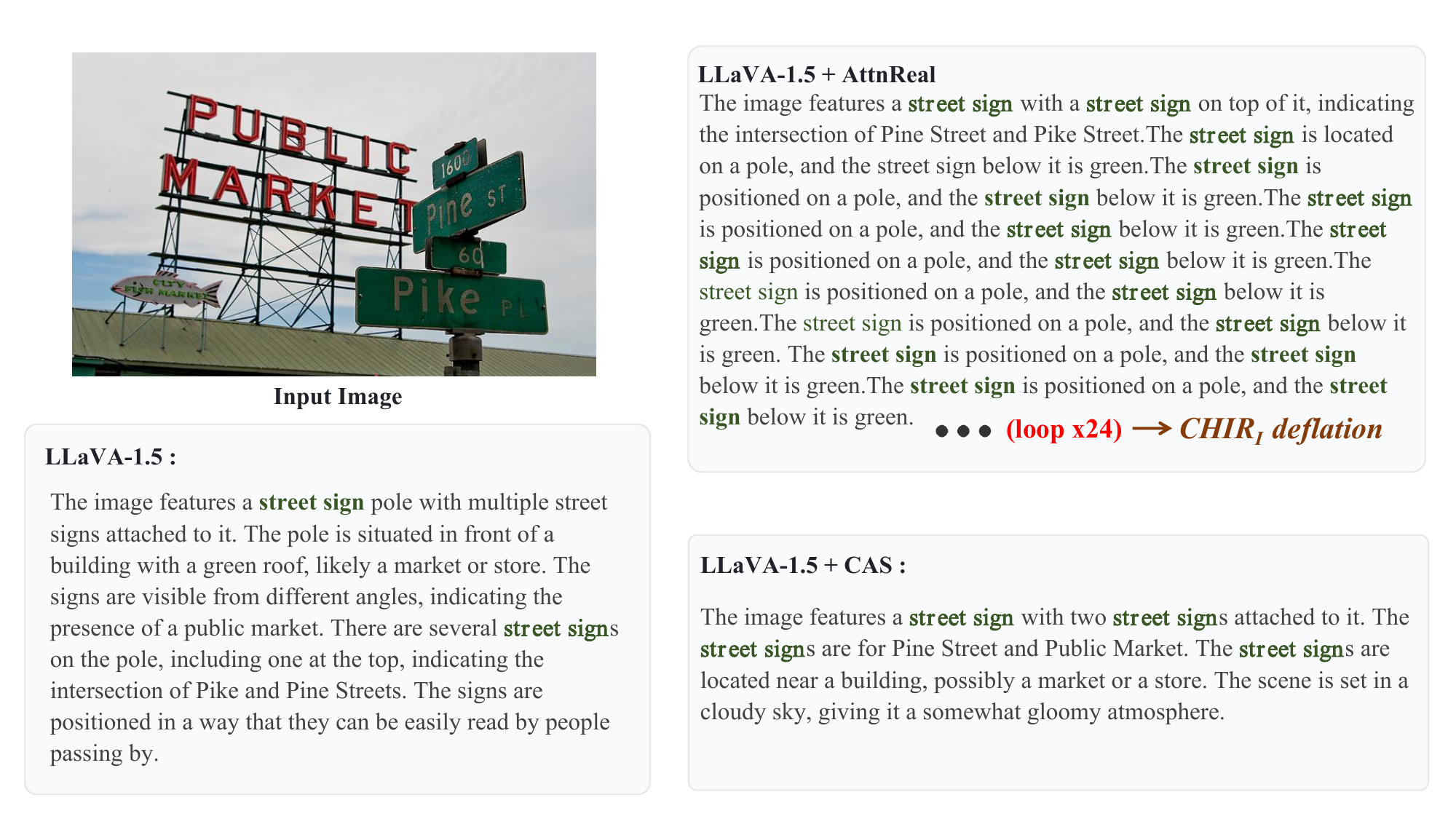}
\caption{\textbf{Repetitive degeneration of AttnReal.} The same CHAIR$_I$ deflation pattern as Figure~\ref{fig:case_pai}: repetition inflates the denominator and depresses the metric reading, while the caption itself has failed. CAS produces a normal generation on the same image.}
\label{fig:case_attnreal}
\end{figure*}

We present two representative cases (Figure~\ref{fig:case_pai} and Figure~\ref{fig:case_attnreal}) showing how strong attention-based interventions (PAI and AttnReal) push the base MLLM into severe repetitive degeneration. The caption contains very few hallucinated objects, but is paired with a multiplied repetition of GT-grounded objects. Because CHAIR$_I$ is counted at the token-instance level without deduplication, the inflated denominator from repetition combined with a sparse hallucination numerator produces a mechanical imbalance: CHAIR$_I$ reads artificially low and masks the underlying generation failure. On the same image, the CAS output (rightmost panel of each figure) maintains vanilla-level fluency without introducing repetition.

\end{document}